\documentclass[journal]{IEEEtran}
\usepackage{times}
\usepackage{epsfig}
\usepackage{graphicx}
\usepackage{amsmath}
\usepackage{amssymb}
\usepackage{multirow}
\usepackage{bm}
\usepackage{changepage}
\usepackage{float}
\usepackage{color}
\usepackage{array}
\usepackage{cite}
\usepackage{fancyhdr}
\usepackage{lipsum}
\usepackage[colorlinks,linkcolor=red]{hyperref}

\begin{document}
	
	%
	\title{A Mutual Bootstrapping Model for Automated Skin Lesion Segmentation and Classification}
	%
	%
	%
	
	\author{Yutong~Xie$\dagger$,
	Jianpeng~Zhang$\dagger$, 
	Yong~Xia, 
	and~Chunhua~Shen
	\thanks{This work was supported in part by the National Natural Science Foundation of China under Grants 61771397, in part by the Science and Technology Innovation Committee of Shenzhen Municipality, China, under Grants JCYJ20180306171334997, and in part by Innovation Foundation for Doctor Dissertation of Northwestern Polytechnical University. 
		({\em Corresponding author: Y. Xia})}.
	\thanks{Y. Xie, J. Zhang, and Y. Xia are with the National Engineering Laboratory for Integrated Aero-Space-Ground-Ocean Big Data Application Technology, School of Computer Science and Engineering, Northwestern Polytechnical University, Xi'an 710072, China (e-mail: xuyongxie@mail.nwpu.edu.cn; james.zhang@mail.nwpu.edu.cn; yxia@nwpu.edu.cn).
		C. Shen is with the School of Computer Science, University of Adelaide, SA, Australia (chunhua.shen@adelaide.edu.au). $\dagger$ indicates equal contribution. The first two authors' contribution was made partly when visiting The University of Adelaide.}
}
	
	%
	%

	\markboth{Accepted at IEEE Transactions on Medical Imaging}
	{Xie \MakeLowercase{\textit{et al.}}: A Mutual Bootstrapping Model for Automated Skin Lesion Segmentation and Classification}
	%
	
	\makeatletter
	\def\ps@IEEEtitlepagestyle{
		\def\@oddfoot{\mycopyrightnotice}
		\def\@evenfoot{}
	}
	\def\mycopyrightnotice{
		{\footnotesize
			\begin{minipage}{\textwidth}
				\centering
				Copyright (c) 2019 IEEE. Personal use of this material is permitted. However, permission to use this material for any other purposes must be obtained from the IEEE by sending a request to pubs-permissions@ieee.org.
			\end{minipage}
		}
	}

	\maketitle
	

	\begin{abstract}
		Automated skin lesion segmentation and classification are two most essential and related tasks in the computer-aided diagnosis of skin cancer. Despite their prevalence, deep learning models are usually designed for only one task, ignoring the potential benefits in jointly performing both tasks. In this paper, we propose the mutual bootstrapping deep convolutional neural networks (MB-DCNN) model for simultaneous skin lesion segmentation and classification. This model consists of a coarse segmentation network (coarse-SN), a mask-guided classification network (mask-CN), and an enhanced segmentation network (enhanced-SN). On one hand, the coarse-SN generates coarse lesion masks that provide a prior bootstrapping for mask-CN to help it locate and classify skin lesions accurately. On the other hand, the lesion localization maps produced by mask-CN are then fed into enhanced–SN, aiming to transfer the localization information learned by mask-CN to enhanced-SN for accurate lesion segmentation. In this way, both segmentation and classification networks mutually transfer knowledge between each other and facilitate each other in a bootstrapping way. Meanwhile, we also design a novel rank loss and jointly use it with the Dice loss in segmentation networks to address the issues caused by class imbalance and hard-easy pixel imbalance. We evaluate the proposed MB-DCNN model on the ISIC-2017 and PH2 datasets, and achieve a Jaccard index of 80.4\% and 89.4\% in skin lesion segmentation and an average AUC of 93.8\% and 97.7\% in skin lesion classification, which are superior to the performance of representative state-of-the-art skin lesion segmentation and classification methods. Our results suggest that it is possible to boost the performance of skin lesion segmentation and classification simultaneously via training a unified model to perform both tasks in a mutual bootstrapping way.
	\end{abstract}
	

	%
	\IEEEpeerreviewmaketitle

	\vspace{-0.1cm}
	\section{Introduction}
	%
	%
	%
	%
	\IEEEPARstart{S}{kin} cancer is one of the most common malignancies to affect the elderly worldwide~\cite{1,2}. Dermoscopy is one of the essential means to improve the diagnostic performance and reduce skin cancer deaths~\cite{3}. Dermoscopic images produced globally are currently analyzed by dermatologists almost entirely through visual inspection, which requires a high degree of skill and concentration, and is time-consuming and prone to operator bias. Computer-aided diagnosis (CAD) has been increasingly studied to assist dermatologists in not only bypassing these issues but also improving the accuracy, efficiency and objectivity of the diagnosis.
	
	There are two significant tasks in building a CAD system of skin cancer: skin lesion segmentation and classification~\cite{4,17}. The segmentation task is used to detect the locations and boundaries of lesions, whereas the classification task is used to diagnose the types of them (e.g. melanoma, nevus, seborrheic keratosis, etc.). Both tasks are challenging due to three reasons: (1) the low contrast between each lesion and its surrounding skin tissue results in fuzzy lesion boundaries; (2) the inter-type skin lesions may share visual similarities, and the intra-type lesions may have visual differences; and (3) skin lesions vary significantly in the visual appearance, which may be corrupted by artifacts like hair, blood vessels, and air bubbles.
	
	A mass of automated skin lesion segmentation and classification methods have been proposed in the literature~\cite{4,5,6,7,8,9,10,11,12,13,14,15,16,17,18,19,20,21,22,23,24,25,26,27,28,29}. Among them, those based on deep convolutional neural networks (DCNNs) have achieved remarkable success ~\cite{4,5,6,7,8,9,18,19,20,21,22,23,28,29}, which are usually designed for either segmentation or classification task. However, skin lesion segmentation and classification are two highly related tasks. The segmentation can help remove distractions from dermoscopic images, and thus is highly beneficial for improving the accuracy of lesion classification, while class-specific diagnostic information can help highlight the lesion regions, and hence contributes to lesion segmentation. Inspired by this, Yu et al.~\cite{24} and {\color{black}González-Díaz}~\cite{26} incorporated lesion segmentation into the diagnosis process, using the segmentation results to filter the distractions and improve the classification. However, they did not explore the potential contribution of skin lesion classification to the segmentation task. 
	
	The benefit of segmentation to classification is obvious, since the segmentation results provide the region of interests (ROIs), in which the discriminative features can be extracted. However, the classification process relies on and produces only the information of image-level class labels. To use such information for image segmentation, we need employ weakly supervised learning (WSL) methods~\cite{30,31,32,33,34}, which usually consist of three successive steps: (1) leveraging a classification model to obtain a localization map of the target, (2) creating proxy pixel-level labels for image-level labeled images based on the localization map, and (3) using proxy pixel-level labels to train a segmentation model. Nevertheless, a classification model focuses only on the class-specific part of an image, which is not necessarily the entire target, particularly in fine-grained type classification. Hence, the localization map produced by the classification model may contain part-lesion and non-lesion regions and results very likely in less-accurate proxy labels, which provide bad supervision to the segmentation task and thereby lead to unsatisfying segmentation results.
	
	In this paper, we propose the mutual bootstrapping DCNNs (MB-DCNN) model for simultaneous skin lesion segmentation and classification. This model consists of a coarse segmentation network (coarse-SN), a mask-guided classification network (mask-CN), and an enhanced segmentation network (enhanced-SN). Both coarse-SN and enhanced-SN are trained on data with pixel-level labels (i.e. dense annotation), and mask-CN is trained on data with image-level class labels. We first employ the lesion mask generated by coarse-SN to boost the lesion localization and discrimination ability of mask-CN, and then transfer the improved localization ability from mask-CN to enhanced-SN for accurate skin lesion segmentation. In this way, both segmentation and classification tasks mutually benefit each other, and their performance improves together. Meanwhile, we use a weighted Dice-and-rank loss for segmentation networks, in which the Dice loss has good compatibility with class-imbalanced pixels and is beneficial for small lesion segmentation, whereas the newly designed rank loss poses additional constraints on hard pixels and thus addresses the issues caused by fuzzy boundaries and artifacts. We have evaluated the proposed model on the benchmark ISIC-2017 dataset~\cite{35} and PH2 dataset~\cite{36}, and achieved more accurate skin lesion segmentation and classification performance than the state-of the-art methods.
	
	The contributions of this work are two-fold. First, based on the idea of leveraging the intrinsic correlation existed in  segmentation and classification tasks, we propose the MB-DCNN model to perform skin lesion segmentation and classification simultaneously and boost the performance of both tasks. Specifically, we not only transfer the lesion masks generated by coarse-SN to mask-CN to provide it the prior lesion location information and improve its ability to extract discriminative representations for lesion classification, but also transfer the lesion location information refined by mask-CN to enhanced-SN to facilitate lesion segmentation, avoiding the troublesome production and use of proxy pixel labels. Second, we propose a novel rank loss and jointly use it with the Dice loss in segmentation networks to enhance the segmentation compatibility on images with class-imbalance and hard-easy pixel imbalance.
	
	\vspace{-0.1cm}
	\section{Related Work}
	Automated skin lesion segmentation and classification have been studied extensively in the literature~\cite{4,5,6,7,8,9,10,11,12,13,14,15,16,17,18,19,20,21,22,23,24,25,26,27,28,29}. Traditionally, skin lesion segmentation methods mainly include clustering, thresholding, region growing, and active contour models~\cite{10,11,12}. Skin lesion classification methods focus mainly on extracting handcrafted features, including the color, texture, border irregularity, and asymmetry descriptors of lesions~\cite{13,14,15,16,17}{~\color{black}and using} one or more of these features to train a classifier such as the K-nearest-neighbor~\cite{17}, back propagation neural network~\cite{13}, support vector machine~\cite{15}, linear classifier~\cite{16}, and logistic regression and product units~\cite{14}. Despite their prevalence, these methods rely heavily on handcrafted features and suffer a lot from less accuracy.

	Since DCNNs offer a unified learning-based solution to feature extraction and classification and avoid the troublesome handcrafted feature extraction, recent years have witnessed the widespread and unprecedented use of DCNNs in medical image analysis~\cite{4,5,6,7,8,9,18,19,20,21,22,23,24,25,26,27,28,29, 29_1, 29_2}. Several DCNN-based skin lesion segmentation methods have been published~\cite{4,5,6,7,8,9,25}. Bi et al.~\cite{6} proposed a multistage segmentation method, in which early stage fully convolutional networks (FCNs) learn appearance and localization features and late stage FCNs learn the subtle characteristics of the lesion boundaries. Yuan et al.~\cite{4} developed a 19-layer deep FCN, which was optimized by using the Jaccard distance loss. Li et al.~\cite{5} presented a new dense deconvolutional network based on the residual learning. Mirikharaji et al.~\cite{9} proposed to encode the star shape prior into the loss function to guarantee a global structure in each segmentation result. Sarkar et al.~\cite{8} presented a robust deep encoder-decoder network to improve the accuracy of obtained lesion boundaries. By contrast, more research efforts have been devoted to the classification of skin lesions~\cite{18,19,20,21,22,23,24,25,26,27,28,29}. Yu et al.~\cite{23} aggregated deep features produced by various layers of a residual network using Fisher vector (FV) encoding. Ge et al.~\cite{22} reported a multi-modality DCNN that accepts both the clinical and dermoscopic views of a single lesion and is capable of learning single-modality and cross-modality representations, simultaneously. Hagerty et al.~\cite{21} combined handcrafted features and the deep features extracted by ResNet-50~\cite{37} for the diagnosis of melanoma. Gessert et al.~\cite{20} proposed a patch-based attention architecture with the diagnosis-guided weighted loss. Zhang et al.~\cite{18,19} proposed the synergic deep learning (SDL) model that uses dual DCNNs to address the issues of intra-class variation and inter-class similarity. Zhang et al.~\cite{29} designed the attention residual learning (ARL) CNN, in which each ARL block jointly uses the residual learning and a novel attention learning mechanism to improve its ability to discriminative representation.
	
	Despite performance improvements, these DCNN-based methods are limited to perform either the segmentation or classification of skin lesions, ignoring the intrinsic correlation between segmentation and classification, which may lead to mutual benefits to both tasks. To address this issue, Yu et al.~\cite{24} introduced a two-stage deep learning framework, under which they first separated the lesion from background, and then cropped the segmented regions as the input of a classification network. Diaz et al. ~\cite{26} further incorporated the extra lesion structure segmentation network into the diagnosis of skin lesions. Both methods utilize the lesion segmentation results to filter the distractions, and thus improve the classification performance. However, they failed to explore the potential benefit of classification results to the lesion segmentation task, which can be achieved by using the WSL strategy.
	
	WSL methods mainly focus on generating high-quality object cues based on image-level labels to supervise the segmentation. Hong et al.~\cite{30} proposed a decoupled network for weakly-supervised segmentation, where the class-specific localization cues are transferred from the classification network to the segmentation network. Kolesnikov et al.~\cite{33} proposed the unified "seed, expand and constrain" (SEC) framework, under which object localization cues, expansion and refining boundary are integrated for segmentation. However, both methods can only provide small and sparse class-related regions to supervise the segmentation, which are not sufficiently accurate for training a reliable segmentation model. To address this drawback, Wei et al.~\cite{38} presented an adversarial erasing method to progressively train multiple classification networks for expanding object regions. Shen et al.~\cite{32} proposed a bi-directional transfer learning framework to generate high quality object localization masks. Yu et al.~\cite{39} and Wei et al.~\cite{31} employed the dilated convolutions to enlarge the receptive field and gain more accurate object regions. Although these methods achieve improvements, they only use the classification network to generate localization maps. Different from these attempts, we reveal that a coarse lesion mask can provide a prior bootstrapping to help the classification network better localize and diagnose lesions. Furthermore, both the decoupled network~\cite{30} and our model feed the class-related localization maps generated by a classification network to a decoder. However, the decoupled network discards encoder and only trains the decoder, where the error accumulation caused by inaccurate localization can hardly be corrected. Our model fuse image features produced by the encoder and the localization maps as the input of the decoder, and then train the encoder and decoder. Hence, our model can ease the impact of inaccurate localization, which is substantial when using only the localization maps.
	
	\begin{figure}[t]
		\centering
		\includegraphics[width=1\linewidth]{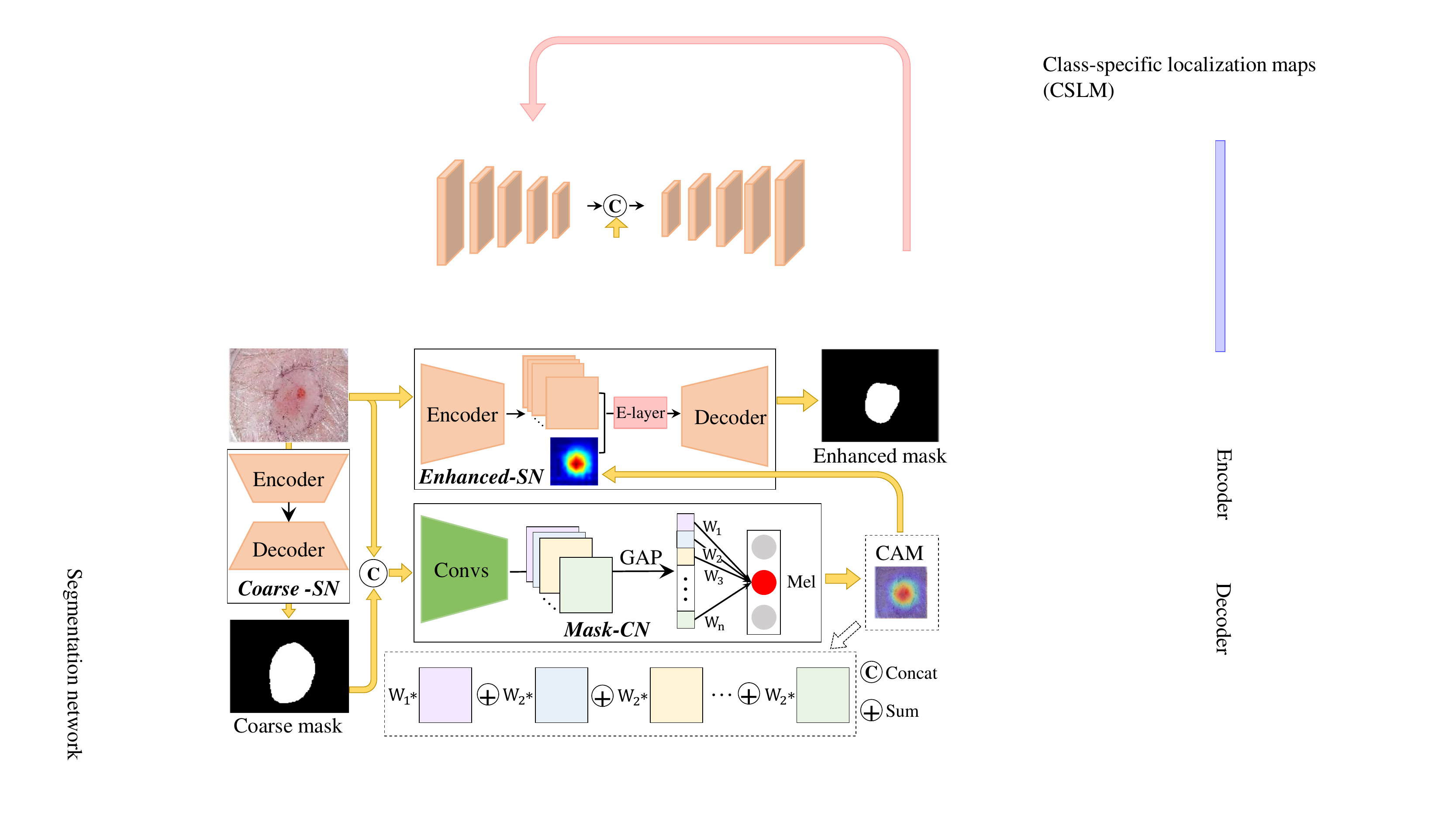}
		\vspace{-0.5cm}
		\caption{Illustration of the proposed MB-DCNN model, which consists of three DCNNs: coarse-SN, mask-CN and enhanced-SN. The coarse-SN is constructed to generate the coarse lesion masks, which are concatenated with original images as the input to train mask-CN. The image features extracted by the encoder of enhanced-SN and the lesion localization maps produced by mask-CN are first fused by an E-layer, and then fed into the decoder of enhanced-SN to generate segmentation results.}
		\label{fig_1}
		\vspace{-0.3cm}
	\end{figure}
	
	\vspace{-0.1cm}
	\section{Method}
	Let the segmentation training set with $N_1$ images be denoted by $I_N={(\bm{X}_n,\bm{Y}_n)}^{N_1}$, in which each image $\bm{X}_n$ is annotated on a pixel-by-pixel basis and each pixel belongs to either the skin lesion (i.e. $y_{ni}$=1) or background (i.e. $y_{ni}$=0). Let the classification training set with $N_2$ ($N_2>N_1$) images be denoted by $I_M={(\bm{X}_m,\bm{Y}_m)}^{N_2}$, in which each image $\bm{X}_m$ is annotated with an image-level label $y_m \in \left \{ l_1,...,l_C \right \}$, and $C$ is the number of classes. The proposed MB-DCNN model consists of three DCNNs: coarse-SN, mask-CN, and enhanced-SN. The pipeline of this model is summarized in Fig.~\ref{fig_1}. First, we train coarse-SN on the dataset $I_N$ for lesion segmentation. Then, we concatenate the images in $I_M$ with the corresponding lesion masks generated by coarse-SN as the input of mask-CN to boost its performance in lesion classification. Finally, we use an enhanced layer (E-layer) to fuse the features of images in $I_N$ extracted by the encoder of enhanced-SN and the corresponding lesion localization maps refined by mask-CN, and then feed them to the decoder of enhanced-SN for more accurate lesion segmentation. In this way, both segmentation and classification networks mutually transfer knowledge between each other and boost the performance of each other in a bootstrapping way. We now delve into the details of this model.
	
	\vspace{-0.3cm}
	\subsection{Coarse-SN}
	We first train coarse-SN using the dataset $I_N$ to roughly segment skin lesions. The obtained coarse masks provide mask-CN the prior information about lesion locations, and hence can enhance the localization and discrimination ability of mask-CN. This step can be viewed as the initialization of lesion masks. {\color{black} We employ the state-of-the-art semantic segmentation network Deeplabv3+ with the modified aligned Xception as its encoder~\cite{40}, which was pre-trained on the MS-COCO~\cite{41} and PASCAL VOC 2012 datasets~\cite{42}, as the backbone of coarse-SN.}
	To adapt the Deeplabv3+ network to our skin lesion segmentation task, we remove its last convolutional layer, and then {\color{black}add} a new convolutional layer with the output channel of one for prediction. The weights of the new layer are randomly initialized, and the activation function in the last layer is set to the sigmoid function. 
	
	\vspace{-0.3cm}
	\subsection{Mask-CN} 
	
	\begin{figure}[t]
		\centering
		\includegraphics[width=0.6\linewidth]{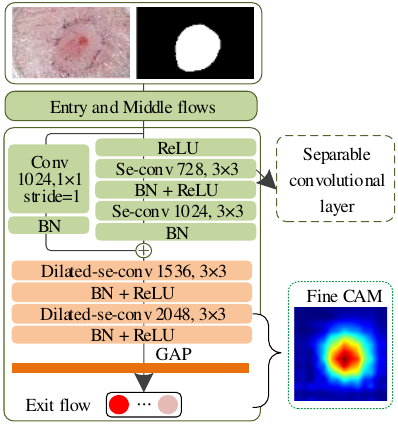}
		\vspace{-0.1cm}
		\caption{Architecture of mask-CN, which is built upon the pre-trained Xception~\cite{43}. The last pooling layer of Xception is replaced by two dilated convolutional layers with a dilated rate of 2.}
		\label{fig_2}
		\vspace{-0.3cm}
	\end{figure}
	
	We use the coarse lesion masks generated by coarse-SN to boost the lesion localization and discrimination ability of the classification network mask-CN, which is trained on the dataset $I_M$, in which each image has only an image-level class label. The region of a skin lesion usually occupies only a small part of a dermoscopic image, and most parts of the image are normal skin tissues with artifacts such as hair, frames, blood vessels, and air bubbles, which may interfere with the lesion classification. The lesion masks can help remove distractions from dermoscopic images, and thus are highly beneficial for accurate discrimination of lesions. 
	
	Each classification training image $\bm{X}_m$ and its corresponding coarse lesion mask are concatenated as an input to mask-CN, whose architecture is shown in Fig.~\ref{fig_2}. This network is built upon the popular classification network Xception~\cite{43}, which was pre-trained on the ImageNet dataset~\cite{44}. We first remove the last pooling layer of Xception to enlarge the resolution of feature maps and avoid the loss of information about small lesions in progressive {\color{black}down-sampling}. Then, we replace the last two separable convolutions of Xception with the separable dilated convolution with a dilated rate of 2, aiming to compensate for the reduction in receptive field caused by removing {\color{black}down-sampling}. After performing the global average pooling (GAP), the produced features are fed to a new fully connected (FC) layer with $C$ neurons, followed by a softmax activation function. The weights of the FC layer are randomly initialized. The weights of the 4th channel (i.e., coarse mask) on input are initialized by averaging the weights of another three channels (i.e., RGB image). We optimize the mask-CN by minimizing the cross-entropy loss. 
	
	\begin{figure}[t]
		\centering
		\includegraphics[width=1\linewidth]{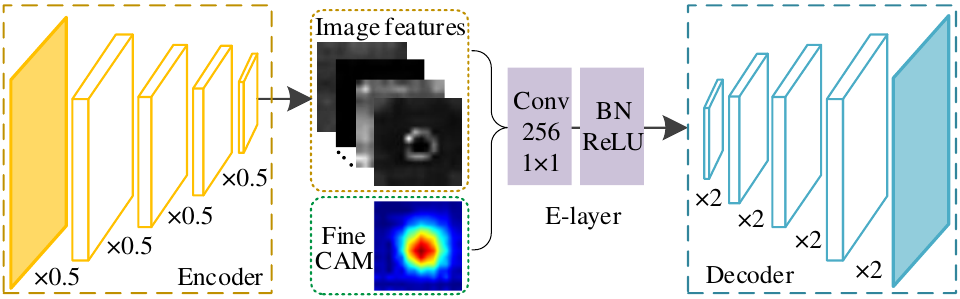}
		\vspace{-0.5cm}
		\caption{Architecture of enhanced-SN which consists of an encoder, an E-layer, and a decoder. The encoder and decoder have the same architecture with those in coarse-SN, and the learnable E-layer is deigned to fuse image features extracted by the encoder and the fine CAM produced by mask-CN.}
		\label{fig_3}
		\vspace{-0.3cm}
	\end{figure}
	
	\vspace{-0.3cm}
	\subsection{Enhanced-SN} 
	For each segmentation training image $\bm{X}_n$, we feed it into the trained mask-CN and use the output of the last convolutional layer to produce the class-specific localization maps $\bm{M}_n$ via the classification activation mapping (CAM)~\cite{45}. Specifically, we first weight the feature maps produced by the last convolutional layer of mask-CN using the class-specific weights of the output layer, and then sum up the weighted feature maps over all channels to generate the class-specific localization maps $\bm{M}_n$. This process is illustrated in Fig.~\ref{fig_1}.
	
	The architecture of enhanced-SN is shown in Fig.~\ref{fig_3}, which consists of an encoder, a decoder, and an E-layer between them. The encoder and decoder share the same architecture and parameters with coarse-SN. To incorporate the refined prior information about lesion locations in $\bm{M}_n$ into the segmentation process, the E-layer first concatenates the feature maps produced by the encoder with $\bm{M}_n$, and then employs a $1\times1$ convolutional layer followed by a batch normalization (BN) layer and the ReLU activation function for information fusion, i.e. fusing the high-level image features and lesion localization information. The resultant feature maps produced by the E-layer are fed into the decoder for fine segmentation. We randomly initialize the weights of the E-layer and train the enhanced-SN using pixel-level labels.
	
	\vspace{-0.2cm}
	\subsection{Hybrid loss for coarse- and enhanced-SN}
	To optimize both coarse-SN and enhanced-SN, we propose the following hybrid loss function
	\begin{equation}
	L_{hybrid}=L_{dice}+\lambda L_{rank}
	\end{equation}
	where $L_{dice}$ is the Dice loss, $L_{rank}$ is a rank loss, and {\color{black}$\lambda$ is a weighting factor that controls the contribution of $L_{rank}$.}
	
	The Dice loss measures the degree of agreement between the prediction and ground truth, shown as follows
	\begin{equation}
	\mathcal{L}_{dice}=1-\frac{2\sum_{i=1}^{V}{p}_{i}{y}_{i}}{\sum_{i=1}^{V}({p}_{i}+{y}_{i})+ \varepsilon} 
	\end{equation}
	where $V$ denotes the number of pixels, $p_i$ represents the predicted probability of the $i$th pixel belonging to the lesion, $y_i$ represents the ground truth label of the $i$th pixel, and $\varepsilon$ is a smooth factor. It is commonly recognized that the pixels that can be easily recognized, such as those located in internal lesion and background, contribute little to the optimization, whereas hard pixels (e.g., boundary pixels) provide more information for the learning process. Due to the severe imbalance between hard and easy pixels in each image, it is challenging to train the segmentation network and achieve a high-quality score using the Dice loss alone, although the Dice loss has a strong compatibility on the data with class imbalance~\cite{46}. To tackle this issue, we employ a rank loss to pose additional constraints on hard and easy pixels.
	
	We design an online rank scheme to select hard pixels dynamically based on the prediction error, which origins from the observation that hard pixels usually produce bigger errors than easy pixels. Specifically, we rank the pixels of the lesion and background, respectively, by their error after the forward propagation of each batch. The top $K$ pixels with the largest error in lesion or background are selected as hard pixels in this area. Let $\mathbf{\boldsymbol{H}}_{ni}^0$ and $\mathbf{\boldsymbol{H}}_{nj}^1$ be prediction values of the $i$th hard pixel of background and the $j$th hard pixel of lesion for the $n$th input image. We calculate the rank loss as follows
	\begin{equation}
	\begin{split}
	L_{rank}\left (\mathbf{\boldsymbol{X}}_{n}, \mathbf{\boldsymbol{Y}}_{n} \right ) = \frac{1}{K^2}\sum_{i=1}^{K}\sum_{j=1}^{K} \max \{ 0,\mathbf{\boldsymbol{H}}_{ni}^0(\mathbf{\boldsymbol{X}}_{n}, \mathbf{\boldsymbol{Y}}_{n} )\\-\mathbf{\boldsymbol{H}}_{nj}^1(\mathbf{\boldsymbol{X}}_{n}, \mathbf{\boldsymbol{Y}}_{n})+margin\}
	\end{split}
	\end{equation}
	which enforces $\mathbf{\boldsymbol{H}}_{nj}^1>\mathbf{\boldsymbol{H}}_{ni}^0+margin$ in the training stage. Such a design enables a segmentation network to pay more attention to those hard pixels and thus learn more discriminative information.
	
	\vspace{-0.2cm}
	\section{Datasets}
	\noindent
	Three dermoscopic image datasets were used for this study.
	\noindent
	\textbf{ISIC-2017 dataset.} The 2017 International Skin Imaging Collaboration (ISIC) skin lesion segmentation challenge dataset~\cite{47} contains 2000 training, 150 validation, and 600 test dermoscopic images. Each image is paired with the expert manual tracing of skin lesion boundaries for the segmentation task and the lesion gold standard diagnosis (i.e., melanoma, nevus and seborrheic keratosis) for the classification task. 
	
	\noindent
	\textbf{ISIC additional dataset.} We collected additional 1320 dermoscopic images from the ISIC archive\footnote {\url{https://www.isic-archive.com}} to expand the training set for the classification task. These images have only image-level labels, including 466 melanoma, 32 seborrheic keratosis, and 822 nevus cases.
	
	\noindent
	\textbf{PH2 dataset.} The PH2 public dataset~\cite{36} contains 200 dermoscopic images, including 160 nevus, and 40 melanomas. All of them were obtained on the same conditions through Tuebinger Mole Analyzer system using a 20-fold magnification and paired with the expert manual tracing of the skin lesion boundaries.
	
	\vspace{-0.1cm}
	\section{Experiments and Results}
	\subsection{Implementation details}
	\noindent
	\textbf{Training phase.} In the proposed MB-DCNN model, both coarse-SN and enhanced-SN were trained on the ISIC-2017 training dataset using pixel-level labels, and mask-CN was trained on both the ISIC-2017 training dataset and ISIC additional dataset using image-level labels only. To further enlarge the training dataset, we employed the online data {\color{black}augmentation}, which includes randomly cropping from the central of each training image with the scale from 50\% to 100\% of the original image size, random rotation from 0 to 10 degrees, shear from 0 to 0.1 radian, shift from 0 to 20 pixels, zoom 110\% of width and height, whitening, horizontal and vertical flips. The augmented patches were then resized to {\color{black}$224\times224$} for training. The Adam algorithm~\cite{48} with a batch size of 16 and 32 were adopted to optimize the segmentation and classification networks, respectively. We set the initialized learning rate to 0.0001 and the maximum epoch number to 500, and set the hyper-parameters in the hybrid loss as $\lambda$=0.05, $K$=30, $margin$=0.3. We used the ISIC-2017 validation set to monitor the performance of each network and terminated the training process when a network falls into overfitting.
	
	\noindent
	\textbf{Testing phase.} On the ISIC-2017 dataset, the trained MB-DCNN was directly applied to the ISIC-2017 testing set for skin lesion segmentation and classification. On the PH2 dataset, we performed two experiments: (1) directly testing the MB-DCNN trained on the ISIC-2017 and ISIC additional datasets (i.e. without fine-tuning) on the entire dataset, and (2) performing the four-fold cross-validation, i.e. regarding the trained MB-DCNN as a pre-trained one, using three folds of PH2 data to fine-tune the model, and testing the fine-tuned model on the other fold of PH2 data. In the second experiment, we also adopted the ISIC-2017 validation set to monitor the fine-tuning process to prevent overfitting.
	
	\vspace{-0.3cm}
	\subsection{Evaluation metrics}
	We evaluated the obtained segmentation results using five performance metrics suggested by the ISIC-2017 challenge, including the Jaccard index (JA), Dice coefficient (DI), pixel-wise accuracy (pixel-AC), pixel-wise sensitivity (pixel-SE), and pixel-wise specificity (pixel-SP). We evaluated the obtained classification results using four performance metrics, including the area under receive operation curve (AUC), accuracy (AC), sensitivity (SE), and specificity (SP). Note that the ISIC-2017 challenge uses JA to rank the segmentation performance of each method on the test dataset, and uses the average AUC of melanoma classification (i.e. melanoma vs. others) and seborrheic keratosis classification (i.e. seborrheic keratosis vs. others) to rank the classification performance.
	
	\begin{table*}[ht]
		\small
		\caption{Skin lesion segmentation performance of our MB-DCNN and several recent methods on the ISIC-2017 testing and PH2 dataset. }
		\vspace{-0.4cm}
		\begin{center}
			\begin{tabular}{m{1.2cm}<{\centering}|m{1cm}<{\centering}|m{1cm}<{\centering}|m{1.2cm}<{\centering}|m{1.1cm}<{\centering}|m{0.8cm}<{\centering}|m{1.1cm}<{\centering}|m{1.2cm}<{\centering}|m{1.2cm}<{\centering}|m{0.8cm}<{\centering}|m{2.4cm}<{\centering}}
				\hline
				Datasets & \multicolumn{5}{c|}{ISIC-2017} & \multicolumn{5}{c}{PH2} \\ \hline
				Methods & \begin{tabular}[c]{@{}c@{}}CDNN,\\ 2017~\cite{7}\end{tabular} & \begin{tabular}[c]{@{}c@{}}DDN,\\ 2017~\cite{5}\end{tabular} & \begin{tabular}[c]{@{}c@{}}FCN+SSP, \\ 2018~\cite{9}\end{tabular} & \begin{tabular}[c]{@{}c@{}}SLSDeep,\\ 2018~\cite{8}\end{tabular} & \begin{tabular}[c]{@{}c@{}}\textbf{Ours}\\ \end{tabular} & \begin{tabular}[c]{@{}c@{}}mFCNPI, \\ 2017~\cite{6}\end{tabular} & \begin{tabular}[c]{@{}c@{}}RFCN,\\2017~\cite{4}\end{tabular} & \begin{tabular}[c]{@{}c@{}}SLIC,\\2018~\cite{10}\end{tabular} & \begin{tabular}[c]{@{}c@{}}Ours\\ \end{tabular} & \begin{tabular}[c]{@{}c@{}}\textbf{Ours (Fine-tuned)}\\ \end{tabular} \\ \hline
				JA & 76.5 & 76.5 & 77.3 & 78.2 & \textbf{80.4} & 84.0 & - & - & 86.7 & \textbf{89.4} \\ \hline
				DI & 84.9 & 86.6 & 85.7 & 87.8 & \textbf{87.8} & 90.7 & 93.8 & - & 92.6 & \textbf{94.2} \\ \hline
				pixel-AC & 93.4 & 93.9 & 93.8 & 93.6 & \textbf{94.7} & 94.2 & - & 90.4 & 95.8 & \textbf{96.5} \\
				\hline
				pixel-SE & 82.5 & 82.5 & 85.5 & 81.6 & \textbf{87.4} & 94.9 & - & 91.0 & \textbf{97.9} & 96.7 \\ \hline
				pixel-SP & 97.5 & \textbf{98.4} & 97.3 & 98.3 & 96.8 & 94.0 & - & 89.7 & \textbf{95.1} & 94.6 \\ \hline
			\end{tabular}
		\end{center}
		\label{tab:tab1}
		\vspace{-0.3cm}
	\end{table*}
	
	\begin{figure}[t]
		\centering
		\includegraphics[width=1\linewidth]{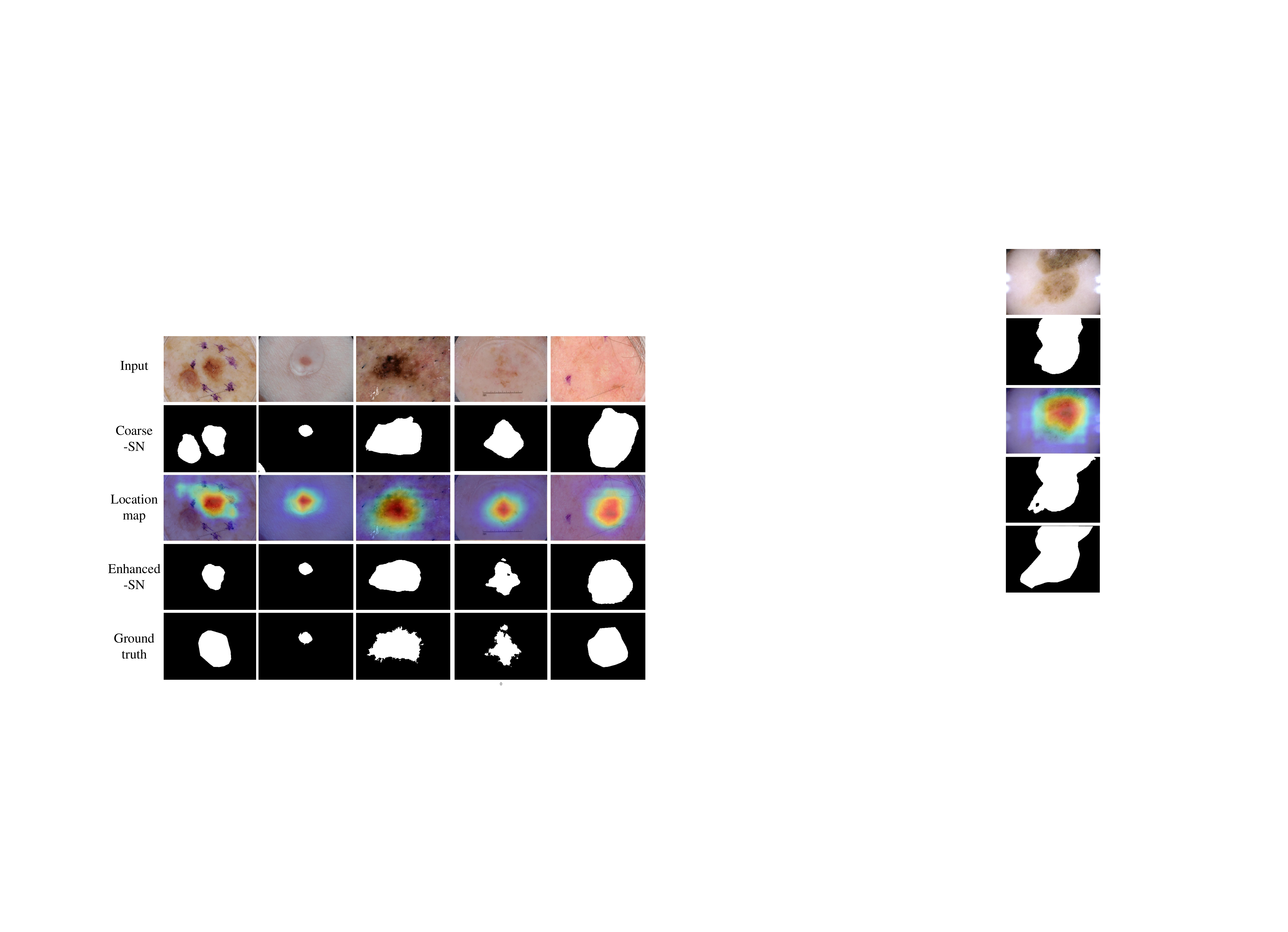}
		\vspace{-0.4cm}
		\caption{Examples of predicted lesion masks by our MB-DCNN model. It shows that (1) the coarse-SN can boost mask-CN to produce more accurate localization maps, and (2) the class-specific localization maps are useful to guide enhanced-SN for achieving better segmentation results.}
		\label{fig_4}
		\vspace{-0.3cm}
	\end{figure}
	
	\vspace{-0.3cm}
	\subsection{Segmentation results}
	We compared the proposed MB-DCNN model to several recently published skin lesion segmentation methods in Table~\ref{tab:tab1}. On the ISIC-2017 dataset, the competing methods include a convolutional-deconvolutional neural network (CDNN)~\cite{7}, a new dense deconvolutional network (DDN)~\cite{5}, a fully convolutional network with star shape prior (FCN+SSP)~\cite{9}, and a skin lesion segmentation deep model based on dilated residual and pyramid pooling network (SLSDeep)~\cite{8}. On the PH2 dataset, the competing methods consist of multi-stage FCN with parallel integration (mFCNPI)~\cite{6}, a retrained FCN (RFCN)~\cite{4}, and a simple linear iterative clustering (SLIC) method~\cite{10}. Note that the performance of all competing methods was adopted in the literature for a fair comparison. 
	
	It shows in Table~\ref{tab:tab1} that our MB-DCNN model achieves the best performance on both datasets. On the ISIC-2017 dataset, our model improves JA by 2.2\% when comparing to the SLSDeep~\cite{8}, which performs the second best. On the PH2 dataset, our model improves JA from 84.0\%, which was achieved by the mFCNPI~\cite{6}, to 86.7\%, although using the models trained on the ISIC-2017 training set and ISIC additional dataset without fine-tuning. {\color{black}These results prove} that our model has a strong generalization ability. Moreover, after fine-tuning the trained models on the PH2 dataset, our model further increases JA to 89.4\% and achieves the best DI of 94.2\%. We also visualizes some segmentation results in Fig.~\ref{fig_4}.
	
	\begin{table*}[ht]
		\small
		\caption{Skin lesion classification performance of our MB-DCNN model, three recent methods, and five top-ranking challenge solutions on the ISIC-2017 testing set.}
		\vspace{-0.4cm}
		\begin{center}
			\begin{tabular}{m{3cm}<{\centering}|m{0.6cm}<{\centering}|m{0.6cm}<{\centering}|m{0.6cm}<{\centering}|m{0.6cm}<{\centering}|m{0.6cm}<{\centering}|m{0.6cm}<{\centering}|m{0.6cm}<{\centering}|m{0.6cm}<{\centering}|m{1.5cm}<{\centering}}
				\hline
				\multirow{2}{*}{Methods} & \multicolumn{4}{l|}{Melanoma Classification} & \multicolumn{4}{l|}{Keratosis Classification} & Average \\ \cline{2-10} 
				& AC & SE & SP & AUC & AC & SE & SP & AUC & AUC (\%) \\ \hline
				\textbf{Ours} & 87.8 & 72.7 & 91.5 & 90.3 & 93.0 & 84.4 & 94.5 & 97.3 & \textbf{93.8} \\ \hline
				ARL-CNN~\cite{29}, 2019 & 85.0 & 65.8 & 89.6 & 87.5 & 86.8 & 87.8 & 86.7 & 95.8 & 91.7 \\ \hline
				SSAC~\cite{49}, 2019 & 83.5 & 55.6 & 90.3 & 87.3 & 91.2 & 88.9 & 91.6 & 95.9 & 91.6 \\ \hline
				SDL~\cite{19}, 2019 & 88.8 & - & - & 86.8 & 92.5 & - & - & 95.8 & 91.3 \\ \hline
				\#1~\cite{50} & 82.8 & 73.5 & 85.1 & 86.8 & 80.3 & 97.8 & 77.3 & 95.3 & 91.1 \\ \hline
				\#2~\cite{51} & 82.3 & 10.3 & 99.8 & 85.6 & 87.5 & 17.8 & 99.8 & 96.5 & 91.0 \\ \hline
				\#3~\cite{52} & 87.2 & 54.7 & 95.0 & 87.4 & 89.5 & 35.6 & 99.0 & 94.3 & 90.8 \\ \hline
				\#4~\cite{53} & 85.8 & 42.7 & 96.3 & 87.0 & 91.8 & 58.9 & 97.6 & 92.1 & 89.6 \\ \hline
				\#5~\cite{54} & 83.0 & 43.6 & 92.5 & 83.0 & 91.7 & 70.0 & 99.5 & 94.2 & 88.6 \\ \hline
			\end{tabular}
		\end{center}
		\label{tab:tab2}
		\vspace{-0.5cm}
	\end{table*}
	
	\begin{table}[ht]
		\small
		\caption{Melanoma and nevus classification performance of our MB-DCNN model and three recent methods on the PH2 dataset.}
		\vspace{-0.4cm}
		\begin{center}
			\begin{tabular}{m{2.5cm}<{\centering}|m{1cm}<{\centering}|m{1cm}<{\centering}|m{1cm}<{\centering}|m{1cm}<{\centering}}
				\hline
				Methods & AC & SE & SP & AUC \\ \hline
				\textbf{Ours (Fine-tuned)}& 94.0 & 95.0 & 93.8 & \textbf{97.7} \\ \hline
				\textbf{Ours} & 88.5 & 82.5 & 90.0 & 95.6 \\ \hline
				CICS~\cite{55}, 2017 & - & 100.0 & 88.2 & - \\ \hline
				MFLF~\cite{56}, 2015 & - & 98.0 & 90.0 & - \\ \hline
				CCS~\cite{57}, 2015 & - & 92.5 & 76.3 & 84.3 \\ \hline
			\end{tabular}
			\label{tab:tab3}
		\end{center}
		\vspace{-0.4cm}
	\end{table}
	
	\vspace{-0.3cm}
	\subsection{Classification results}
	In the meantime, we also compared the proposed MB-DCNN model to several recently published skin lesion classification methods. On the ISIC-2017 dataset, the competing methods include the advanced semi-supervised adversarial classification (SSAC) model~\cite{49}, attention residual learning convolutional neural network (ARL-CNN)~\cite{29}, synergic deep learning (SDL) model~\cite{19}, and five top-ranking methods on the ISIC-2017 skin lesion classification challenge leaderboard~\cite{50,51,52,53,54}. On the PH2 dataset, the competing methods methods are the clinically inspired CAD system (CICS)~\cite{55}, multi-feature late fusion (MFLF)~\cite{56} method, and color constancy system (CCS)~\cite{57}. Similarly, we adopted the performance of all those competing methods in the literature.
	
	The results in Table~\ref{tab:tab2} show that our model achieves the highest average AUC on the ISIC-2017 testing set, not only higher than the top ranking performance on the challenge leaderboard but also 2.1\% higher than the average AUC achieved by the ARL-CNN model~\cite{29}, which, to our knowledge, is the most accurate solution in the literature.
	
	The results in Table~\ref{tab:tab3} show that applying directly our MB-DCNN model, which was trained on the ISIC-2017 training set and ISIC additional dataset without fine-tuning, to the PH2 dataset achieves a noticeably higher AUC than the CCS system~\cite{57} (from 84.3\% to 95.6\%), which proves again that our model has a strong generalization ability. By further fine-tuning the trained MB-DCNN model on the PH2 dataset, our model achieves the highest AUC of 97.7\%, highest AC of 94.0\%, highest SP of 93.8\%, and a comparable SE of 95.0\%. 
	
	\vspace{-0.2cm}
	{\color{black}\subsection{Comparison to other multi-task methods}
	We compared the proposed MB-DCNN model to two popular multi-task methods, which were abbreviated as Method I~\cite{58} and II~\cite{59}, respectively. Method I~\cite{58} shares parameters between the encoder of a segmentation network and the feature extraction (FE) module of a classification network for the joint optimization of both networks. To further utilize the intrinsic correlation between these two tasks, Method II~\cite{59} and our MB-DCNN model propose different learning strategies to transfer the useful information learned on one task to the other task. Method II shares the parameters of lower layers between the encoder and FE module, and introduces the feature passing module at higher layers to transmit the features representation from the encoder to the FE module, and vice versa. This method, however, still shares a pre-defined architecture for different tasks, which limits the flexibility of designing a task-specific network. By contrast, our model uses different networks for different tasks. Moreover, our model first transfers the lesion location produced by a segmentation network to a classification network to improve its lesion classification and localization ability, and then transfers the high-quality localization information learned by the classification network to another segmentation network to facilitate lesion segmentation. Table~\ref{tab:tab4_} gives the performance of our MB-DCNN model and these two multi-task methods on the ISIC-2017 and PH2 datasets. For a fair comparison, the same segmentation network, i.e. Deeplabv3+, was adopted by three methods. It shows that our model achieves the highest segmentation and classification performance metrics on both datasets, which demonstrate the effectiveness of our multi-task strategy.
    }

	\begin{table*}[t]
	{\color{black}{
		\small
		\caption{Skin lesion segmentation and classification performance of our MB-DCNN model and two multi-task methods on the ISIC-2017 and PH2 datasets. Noted that the classification evaluation metrics on ISIC-2017 dataset is calculated by averaging the performance in melanoma classification and seborrheic keratosis classification.}
		\vspace{-0.4cm}
		\begin{center}
		\begin{tabular}{c|c|c|c|c|c|c|c|c|c|c}
			\hline
			\multirow{2}{*}{Data-sets} & Tasks     & \multicolumn{5}{c|}{Segmentation}            & \multicolumn{4}{c}{Classification} \\ \cline{2-11} 
			& Methods   & JA   & DI   & pixel-AC & pixel-SE & pixel-SP & AC      & SE      & SP     & AUC    \\ \hline
			\multirow{3}{*}{ISIC-2017} & Method I  & 77.4 & 85.7 & 93.6     & 85.8     & 95.4     & 87.5    & 72.1    & 91.1   & 91.8   \\ \cline{2-11} 
			& Method II & 78.1 & 86.2 & 94.0     & 86.0     & 95.8     & 89.3    & 75.8    & 92.1   & 92.8   \\ \cline{2-11} 
			& \textbf{Ours}      & \textbf{80.4} & \textbf{87.8} & \textbf{94.7}     & \textbf{87.4}     & \textbf{96.8}     & \textbf{90.4}    & \textbf{78.6}    & \textbf{93.0}   & \textbf{93.8}   \\ \hline
			\multirow{3}{*}{PH2}       & Method I  & 88.0 & 93.3 & 95.8     & 96.6    & 93.7     & 92.0   & 92.5   & 91.9  & 96.6   \\ \cline{2-11} 
			& Method II & 88.6 & 93.7 & 96.1     & 95.7     & \textbf{95.3}     & 92.5   & \textbf{95.0}   & 91.9  & 97.3   \\ \cline{2-11} 
			& \textbf{Ours}      & \textbf{89.4} & \textbf{94.2} & \textbf{96.5}     & \textbf{96.7}     & 94.6    & \textbf{94.0}    & \textbf{95.0}  & \textbf{93.8}   & \textbf{97.7} \\ \hline
		\end{tabular}	
		\end{center}
		\label{tab:tab4_}}}
		\vspace{-0.4cm}
	\end{table*}
	
	\vspace{-0.2cm}
	\subsection{Hybrid loss and its parameter settings}
\begin{figure}[t]
		\centering
		\includegraphics[width=0.9\linewidth]{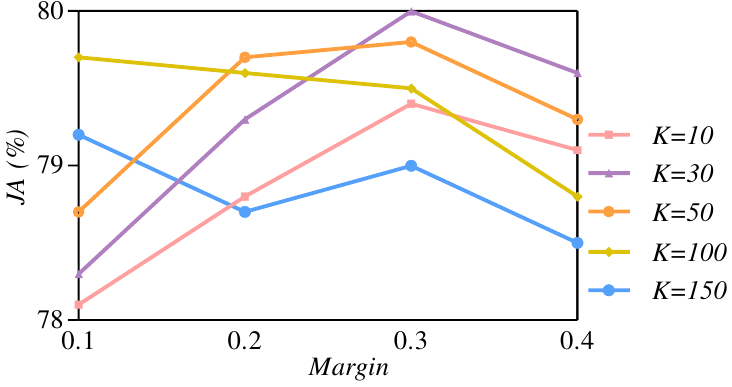}
		\vspace{-0.3cm}
		\caption{JA values obtained on the ISIC-2017 validation set versus the values of $K$ and $margin$.}
		\label{fig_5}
		\vspace{-0.1cm}
	\end{figure}

	{\color{black} In the proposed rank loss, there are two important hyper-parameters, i.e. $K$ and $margin$, which represent the number of selected hard pixels and the constraint between the hard pixels of background and foreground, respectively.} To investigate the impact of their settings on the segmentation, we attempted to set $K$ to 10, 30, 50, 100, and 150, and set $margin$ to 0.1, 0.2, 0.3, and 0.4. We plotted the JA values obtained on the ISIC-2017 validation set versus the values of $K$ and $margin$ in Fig.~\ref{fig_5}. It is clear that the proposed MB-DCNN model achieves the highest JA when setting $K$ is set to 30 and $margin$ is set to 0.3. Hence, we empirically set $K$ to 30 and $margin$ to 0.3 for this study. 
	
\begin{figure}[t]
		\centering
		\includegraphics[width=0.6\linewidth]{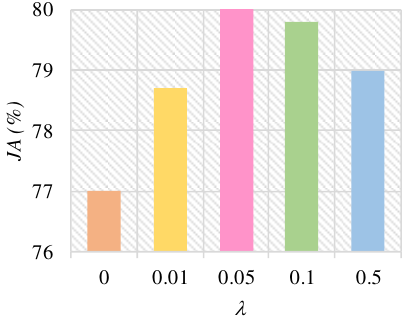}
		\vspace{-0.3cm}
		\caption{JA values obtained on the ISIC-2017 validation set when the weighting factor $\lambda$ is set to different values.}
		\label{fig_6}
		\vspace{-0.3cm}
	\end{figure}

	{\color{black} In the hybrid loss, the weighting factor $\lambda$ is a critical parameter, which controls the contribution of the rank loss. To investigate the setting of this parameter, we fixed the value of $K$ to 30 and $margin$ to 0.3, and repeated the skin segmentation experiment with different values of $\lambda$, including 0.01, 0.05, 0.1, and 0.5. The JA values obtained on the ISIC-2017 validation set is depicted in Fig.~\ref{fig_6}. Obviously, the proposed MB-DCNN model achieves the highest JA when $\lambda$ is set to 0.05. Therefore, we suggest using 0.05 as the default weighting factor in the hybrid loss.} 
	
	{\color{black} To demonstrate the performance gain resulted from the proposed hybrid loss, we also attempted to train our model with different loss functions, including the weighted cross-entropy (WCE), Dice loss, and focal loss~\cite{focal}. The WCE and Dice loss mainly aim to deal with the class-imbalance issue, whereas the focal loss~\cite{focal} and our hybrid loss take into account the imbalance issues related to both classes and hard-easy pixels. Table~\ref{tab:tab4} gives the segmentation performance obtained when using different loss functions. It reveals that (1) when considering only class-imbalance, the Dice loss performs better than the WCE loss; (2) when dealing with both imbalance issues, our hybrid loss achieves better performance than focal loss; and (3) the superior performance of our hybrid loss over the Dice loss confirms the effectiveness of using the rank loss to pose constraints to hard pixels.}

	\begin{table}[t]
		{\color{black}{
		\small
		\caption{Skin lesion segmentation performance of our MB-DCNN when using different loss functions on the ISIC-2017 validation set.}
		\vspace{-0.8cm}
		\begin{center}
			\begin{tabular}{c|m{0.4cm}<{\centering}|m{0.4cm}<{\centering}|m{1.2cm}<{\centering}|m{1.2cm}<{\centering}|m{1.1cm}<{\centering}}
				\hline
				Loss Function & JA & DI & pixel-AC & pixel-SE & pixel-SP \\ \hline
				WCE loss & 76.4 & 84.3 & 94.8 & 82.9 & \textbf{97.4} \\ \hline
				Dice loss & 77.0 & 84.9 & 95.4 & 83.7 & 95.6 \\ \hline
				Focal loss & 77.8 & 85.5 & 95.6 & 84.6 & 97.2	 \\ \hline
				Our hybrid loss & \textbf{80.0} & \textbf{87.9} & \textbf{96.2} & \textbf{87.4} & 95.3 \\ \hline
			\end{tabular}
		\end{center}
		\label{tab:tab4}}}
		\vspace{-0.5cm}
	\end{table}		
	
	\vspace{-0.1cm}
	\section{Discussions}
	\subsection{Using segmentation to boost classification}
\begin{figure}[t]
		\centering
		\includegraphics[width=0.95\linewidth]{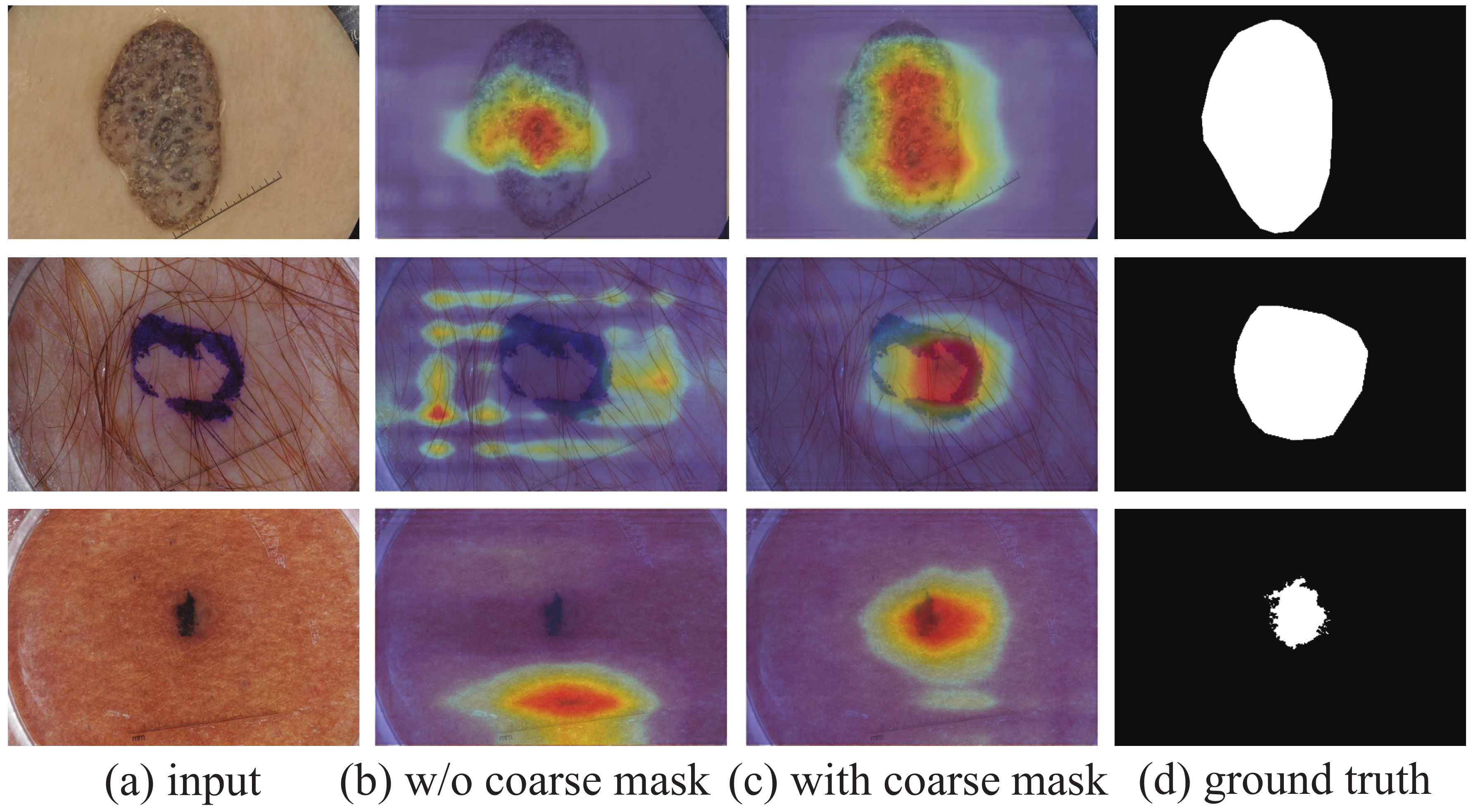}
		\vspace{-0.2cm}
		\caption{Comparison of the CAMs obtained by our MB-DCNN model with or without using the coarse lesion masks: (a) three dermoscopic images, (b) CAMs obtained when not using coarse lesion masks, (c) CAMs obtained when using coarse lesion masks, and (d) ground truth for lesion segmentation.}
		\label{fig_7}
		\vspace{-0.3cm}
	\end{figure}

	In the proposed MB-DCNN model, we concatenate the images and the corresponding lesion masks predicted by coarse-SN as the input of mask-CN, aiming to use the results of skin lesion segmentation to facilitate skin lesion classification. To evaluate the effectiveness of this strategy, we compared the skin lesion classification performance obtained on the ISIC-2017 validation set with or without using the coarse lesion masks produced by coarse-SN in Table~\ref{tab:tab5}. It reveals that, with the bootstrapping of coarse-SN, our model is substantially more accurate in skin lesion classification, improving the average AUC from 94.9\% to 97.0\%. Such performance gain is understandable, since the predicted lesion masks enable mask-CN to focus more on lesions instead of background skin tissues on dermoscopic images and thus strengthen the ability of mask-CN to diagnose different lesion types. To further validate this explanation, we visualized three CAMs obtained by mask-CN with or without using the coarse lesion masks in Fig.~\ref{fig_7}. It shows that, when using the coarse lesion masks, the obtained CAMs are more similar to the segmentation ground truth.  
	
	\begin{table*}[t]
		\small
		\caption{Skin lesion classification performance of our MB-DCNN model with or without (w/o) using the coarse lesion masks on the ISIC-2017 validation set.}
		\vspace{-0.4cm}
		\begin{center}
			\begin{tabular}{c|c|c|c|c|c|c|c|c|c}
				\hline
				\multirow{2}{*}{Methods} & \multicolumn{4}{c|}{Melanoma Classification} & \multicolumn{4}{c|}{Keratosis Classification} & Average \\ \cline{2-10} 
				& AC & SE & SP & AUC & AC & SE & SP & AUC & AUC (\%) \\ \hline
				w/o coarse-SN & 86.7 & 63.3 & 92.5 & 92.6 & 94.0 & 88.1 & 96.3 & 97.1 & 94.9 \\ \hline
				with coarse-SN & 91.3 & 86.7 & 92.5 & 94.6 & 93.3 & 100.0 & 90.7 & 99.4 & \textbf{97.0} \\ \hline
			\end{tabular}
			\label{tab:tab5}
		\end{center}
		\vspace{-0.5cm}
	\end{table*}
	
	\vspace{-0.2cm}
	\subsection{Using classification to boost segmentation}
	A uniqueness of the proposed MB-DCNN model is to transfer the high-quality lesion localization maps learned by a classification network to a segmentation network to facilitate lesion segmentation. Obviously, the quality of lesion localization maps and the manner of transferring the maps affect the segmentation performance.
	
	\noindent
	\textbf{Quality of lesion localization maps.} To assess the impact of the quality of lesion localization maps on the segmentation, we compared the performance of three models on the ISIC-2017 validation set in Table~\ref{tab:tab6}. Model A contains only coarse-SN, which does not use any lesion localization maps. Model B contains both mask-CN and enhanced-SN, but without using coarse-SN. As shown in Fig.~\ref{fig_7}, without the bootstrapping of coarse-SN, the lesion localization maps produced by mask-CN are less accurate. The third model is the proposed one, in which mask-CN provides the high-quality lesion localization maps with the facility of coarse-SN. It shows in Table~\ref{tab:tab6} that transferring localization maps from mask-CN without the bootstrapping of coarse-SN to enhanced-SN results in a lower JA than with the bootstrapping (Model B vs. proposed model), even worse than directly using coarse-SN (Model B vs. Model A). The poor segmentation performance is attributed mainly to the fact that the mask-CN without the bootstrapping of coarse-SN has a weak localization ability that damages the performance of enhanced-SN. Our model uses the mask-CN with the bootstrapping of coarse-SN, and hence can produce high-quality localization maps, which in turn enable enhanced-SN to achieve a better performance than coarse-SN (Proposed model vs. Model A).
	
	\begin{table}[t]
		\small
		\caption{Skin lesion segmentation performance of three models with the lesion localization maps of different quality on the ISIC-2017 validation set.'CS' is coarse-SN. 'MC' is mask-CN. 'ES' is enhanced-SN.}
		\vspace{-0.4cm}
		\begin{center}
			\begin{tabular}{m{0.8cm}<{\centering}|m{0.5cm}<{\centering}|m{0.5cm}<{\centering}|m{0.5cm}<{\centering}|m{0.5cm}<{\centering}|m{0.5cm}<{\centering}|m{0.5cm}<{\centering}|m{0.5cm}<{\centering}|m{0.5cm}<{\centering}}
				\hline
				Models & CS & MC & ES & JA & DI & pixel-AC & pixel-SE & pixel-SP \\ \hline
				A & $\vee$ & $\times$ & $\times$ & 78.9 & 86.5 & 95.7 & 88.7 & 95.0 \\ \hline
				B & $\times$ & $\vee$ & $\vee$ & 77.7 & 86.0 & 95.5 & \textbf{89.1} & \textbf{96.0} \\ \hline
				\textbf{Ours} & $\vee$ & $\vee$ & $\vee$ & \textbf{80.0} & \textbf{87.9} & \textbf{96.2} & 87.4 & 95.3 \\ \hline
			\end{tabular}
			\label{tab:tab6}
		\end{center}
		\vspace{-0.3cm}
	\end{table}
	
	\begin{table}[t]
		\small
		\caption{Skin lesion segmentation performance of three models with different manners of transferring lesion localization maps on the ISIC-2017 validation set.}
		\vspace{-0.8cm}
		\begin{center}
			\begin{tabular}{m{2.25cm}<{\centering}|m{0.4cm}<{\centering}|m{0.4cm}<{\centering}|m{1.2cm}<{\centering}|m{1.15cm}<{\centering}|m{1.1cm}<{\centering}}
				\hline
				Methods & JA & DI & pixel-AC & pixel-SE & pixel-SP  \\ \hline
				D-Net, 2015~\cite{30} & 76.2 & 84.9 & 94.4 & \textbf{89.5} & 94.6  \\ \hline
				MDC, 2018~\cite{31} & 78.9 & 87.0 & 96.1 & 86.1 & \textbf{95.9} \\ \hline
				\textbf{Ours} & \textbf{80.0} & \textbf{87.9} & \textbf{96.2} & 87.4 & 95.3  \\ \hline
			\end{tabular}
			\label{tab:tab7}
		\end{center}
		\vspace{-0.3cm}
	\end{table}
	
	\noindent
	\textbf{Manner of transferring lesion localization maps.} Besides the quality of lesion localization maps, the manner of transferring the maps from mask-CN to enhanced-SN also matters. The MDC method~\cite{31} creates proxy pixel-level labels for the images with only image-level labels based on localization maps, and then trains two segmentation networks by jointly using both real and proxy pixel-level labels. To avoid the impact of inaccurate labels, once localization maps are obtained, the decoupled network (D-Net)~\cite{30} and our MB-DCNN model train only one segmentation network, to which the localization maps are transferred. Nevertheless, the D-Net only transfers localization maps to the decoder, and our model, by contrast, fuses the maps with the image features produced by the encoder and feeds them to the decoder. We compare the segmentation performance of these three models in Table~\ref{tab:tab7}. For a fair comparison, the MDC method and our model use the same segmentation network, i.e. Deeplabv3+~\cite{40}, and the D-Net and our model use the same decoder. It shows that our model is able to ease the error accumulation caused by the inaccurate localization maps, and hence achieves an improvement of 3.8\% in JA over the D-Net. The results also suggest that, with the transfer of lesion localization maps, our model improves JA by 1.1\% over the MDC method.
	
	\vspace{-0.2cm}
	\subsection{Advantages of mutual bootstrapping}
	The proposed MB-DCNN model can not only transfer the lesion location produced by a segmentation network to mask-CN to facilitate lesion classification, but also transfer the high-quality lesion localization information learned by mask-CN to enhanced-SN to alleviate the impact of inaccurate localization on segmentation results. As a result, both segmentation and classification networks mutually facilitate each other in a bootstrapping way, and thus boost the performance of both skin lesion segmentation and classification simultaneously, particularly when the training set is small.
	
\begin{figure}[t]
		\centering
		\includegraphics[width=0.8\linewidth]{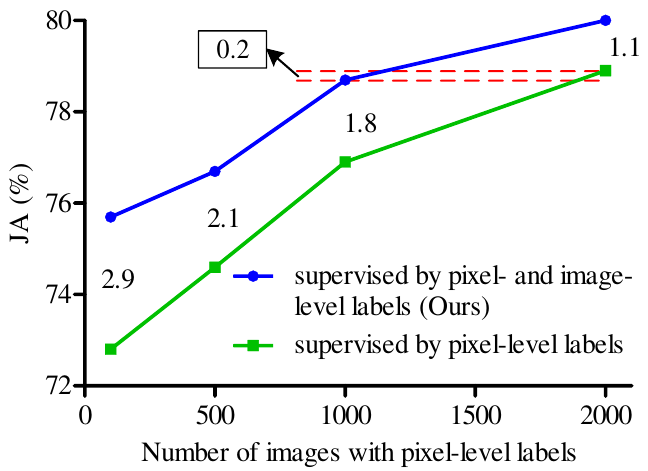}
		\vspace{-0.1cm}
		\caption{JA values achieved on the ISIC-2017 validation set by MB-DCNN model (trained on some images with pixel-level labels and some with image-level labels) and a fully-supervised segmentation network (trained only on images with pixel-level labels), when the number of training images with pixel-level labels increases from 100 to 2000.}
		\label{fig_8}
		\vspace{-0.3cm}
	\end{figure}

	A major advantage of using classification to boost segmentation is to use the images with only image-level annotations to facilitate the training of a segmentation network, leading to less requirement of the pixel-level dense annotation. To demonstrate this advantage, we compared the proposed MB-DCNN model to a fully-supervised segmentation model, which has the same architecture to enhanced-SN and is trained only on the images with pixel-level labels, but without using any lesion localization maps. The JA values of both solutions on the ISIC-2017 validation set versus the number of training images with pixel-level labels were plotted in Fig.~\ref{fig_8}. As expected, with the weak supervision provided by image-level labels, our model outperforms the fully-supervised one steadily no matter how many training images with pixel-level labels were used. However, the performance gain reduces from 2.9\% to 1.1\% when the number of densely annotated training images increases from 100 to 2000. It is not surprising, since the more densely annotated training images are used, the less weakly annotated images are need for auxiliary training. 
	
	More important, it is interesting that the JA achieved by our model trained with 1000 densely annotated images is only 0.2\% less than the JA achieved by the fully supervised method trained with 2000 samples. {\color{black}A similar phenomenon can be found when the number of densely annotated training images reduces to 500 and 1000, respectively.} It suggests that our model provides the possibility of using weakly annotated training images to replace almost half of densely annotated training images while maintaining the segmentation performance.
	
\begin{figure}[t]
		\centering
		\includegraphics[width=0.85\linewidth]{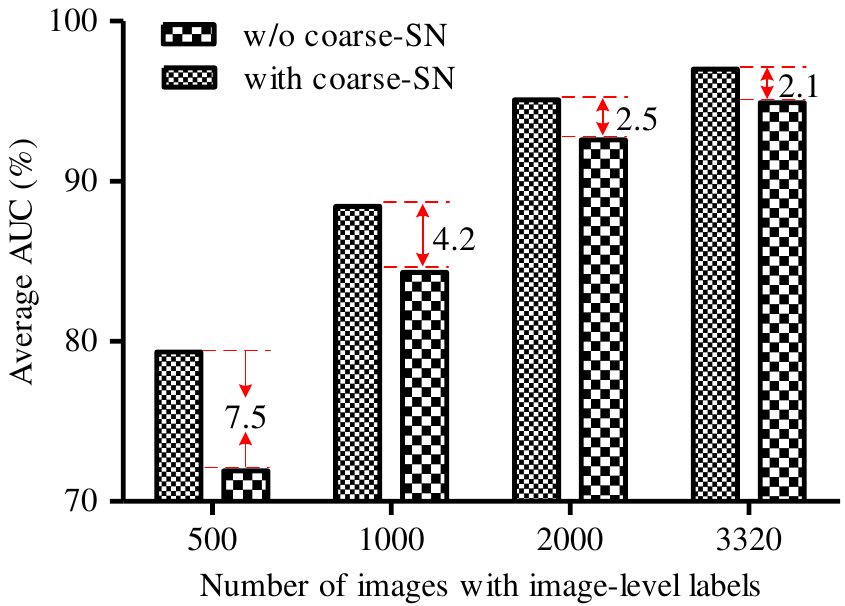}
		\vspace{-0.1cm}
		\caption{Average AUC values obtained on the ISIC-2017 validation set by our MB-DCNN model with or without the help of coarse-SN when the number of image-level annotated training images increases from 500 to 3320.}
		\label{fig_9}
		\vspace{-0.3cm}
	\end{figure}

	By contrast, the advantages of using segmentation to boost classification is quite explicit, since segmentation results provide a classification network the regions of skin lesions, in which discriminative features should be extracted. Although the classification network can generate such regions of interest by itself, using the lesion masks produced by the segmentation network as part of the input results in more accurate regions of interest, as shown in Figure~\ref{fig_7}, which in turn improve the classification performance. To demonstrate the sustainability of such performance improvement, we kept the validation dataset and pixel-level annotated training set unchanged, and selected randomly 500, 1000, and 2000 image-level annotated images to train our MB-DCNN model with or without the help of coarse-SN, respectively. We compared their average AUC on the ISIC-2017 validation set in Figure~\ref{fig_9}. As expected, with the help of the lesion masks, our model outperforms the model without coarse-SN steadily, no matter how many image-level annotated training images were used. However, the classification performance gain reduces from 7.5\% to 2.1\% when the number of training images with image-level labels increases from 500 to 3320. It suggests that the less training data we have, the greater role the lesion masks may play.
	
	Meanwhile, it reveals that, with the help of lesion masks, our model trained on 2000 images has similar performance to the model without using lesion masks but being trained on 3320 images. It indicates that using the lesion masks produced by coarse-SN can largely compensate the loss of  classification power caused by reducing the number of training images.

	\begin{table*}[h]
	{\color{black}{
			\small
			\caption{Skin lesion segmentation and classification performance of our MB-DCNN and six advanced segmentation or classification models on the ISIC archive dataset.}
			\vspace{-0.4cm}
			\begin{center}
				\begin{tabular}{c|c|c|c|c|c|c|c|c|c|c}
					\hline
					Tasks               & \multicolumn{5}{c|}{Segmentation}            &                   & \multicolumn{4}{c}{Classification} \\ \hline
					Methods             & JA   & DI   & pixel-AC & pixel-SE & pixel-SP & Methods           & AC      & SE      & SP     & AUC    \\ \hline
					SLSDeep~\cite{8}    & 75.1 & 84.8 & 96.4     &  \textbf{93.7}     & 95.1     & Xception~\cite{43} & 89.3    & 81.6    & 89.8   & 94.0   \\ \hline
					U-Net~\cite{U-net}      & 79.7 & 87.9 & 97.1     & 91.9     & 96.7     & SDL~\cite{19}      & 90.4    & 83.3    & 90.5   & 94.1   \\ \hline
					DeeplabV3+~\cite{40} & 80.9 & 88.8 & 97.5     & 92.3     & 97.3     & ARL-CNN~\cite{29}  & 90.3    & 83.3    & 90.4   & 94.4   \\ \hline
					Ours                &  \textbf{82.1} &  \textbf{89.7} &  \textbf{97.9}     & 93.3     &  \textbf{97.8}     & Ours              &  \textbf{92.3}    &  \textbf{83.7}    &  \textbf{92.8}   &  \textbf{95.7}   \\ \hline
				\end{tabular}
			\end{center}
			\label{tab:tab8}}}
	\vspace{-0.3cm}
	\end{table*}

	\vspace{-0.2cm}
	{\color{black}\subsection{Robustness to ISIC archive dataset}}
	{\color{black}The ISIC archive dataset\footnote {\url{https://www.isic-archive.com}}  provides more and newer dermoscopic images for skin lesion analysis. In this experiment, we used 10697 dermoscopic images for the segmentation task and 21153 images for the classification task. Each segmentation sample is paired with the expert manual tracing of skin lesion boundaries, and the classification image set includes 18566 nevus, 2168 melanomas, and 419 seborrheic keratosis cases. We evaluated our MB-DCNN model against three advanced segmentation models (i.e. SLSDeep model~\cite{8}, U-Net~\cite{U-net} and DeeplabV3+~\cite{40}) and three classification models (i.e. Xception~\cite{43}, SDL model~\cite{19} and ARL-CNN~\cite{29}) on this dataset using the two-fold cross-validation. For a fair comparison, the performance of all competing models was obtained by running their source code. The obtained lesion segmentation performance and the average performance in melanoma classification and seborrheic keratosis classification were given in Table~\ref{tab:tab8}. It shows that the proposed MB-DCNN model that jointly uses the image-level and pixel-level labeled data for training attains the best performance on both tasks. Specifically, our model improves JA by 1.2\% over the second best model (i.e. DeeplabV3+) on the segmentation task and improves the average AUC from 94.4\% (achieved by ARL-CNN) to 95.7\%. These results further justify the effectiveness and generality of the proposed MB-DCNN model.}
	
	\vspace{+0.3cm}
	\subsection{Complexity}
	For this study, three DCNNs in our MB-DCNN model are trained using the open source Keras and Tensorflow software packages. In our experiments, it took about 48 hours to train our MB-DCNN model (24 hours for coarse-SN, 12 hours for mask-CN and 12 hours for enhanced-SN) and less than 1 second to apply it to segment and classify each lesion on a server with 4 NVIDIA GTX 1080 Ti GPUs and 512GB Memory. Although time-consuming, training the model can be done offline. The fast online testing suggests that our model has the potential to be used in a routine clinical workflow.
	
	{\color{black} Based on the above discussions, we have five suggestions the conclusions: (1) transferring the lesion masks produced by the segmentation network to the classification network can improve the classification performance as well as lesion localization maps; (2) transferring the lesion localization maps produced by the classification network to the segmentation network can alleviate the impact of inaccurate localization and thus improves the lesion segmentation performance; (3) the performance gain caused by the mutual bootstrapping of segmentation and classification is more obvious when the training dataset is even smaller; (4) using the training samples with image-level annotation to facilitate the training of the segmentation model in the proposed way can reduce the requirement of the training samples with pixel-level annotation; and (5) the proposed MB-DCNN model remains effective on the ISIC archive dataset, which contains more and newer dermoscopic images for skin lesion segmentation and classification.}
	
	\vspace{-0.1cm}
	\section{Conclusion}
	In this paper, we have proposed  the MB-DCNN model for simultaneous skin lesion segmentation and classification, which consists of a classification network, i.e. mask-CN, and two segmentation networks, i.e., coarse-SN and enhanced-SN. The coarse-SN provides prior bootstrapping to mask-CN for accurately localization and diagnosis of skin lesions, and then the localization ability is transferred to enhanced-SN to obtain accurate lesion segmentation. We also design a novel hybrid loss to optimize both segmentation networks. Our results on the ISIC 2017 dataset and PH2 dataset not only demonstrate the effectiveness of mutual bootstrapping for image segmentation methods and classification, but also indicate that the proposed MB-DCNN model outperforms the state-of-the-art skin lesion segmentation methods and classification methods with substantial margins. 
	
	{\color{black}
		In the future work, we plan to extend the proposed MB-DCNN model to an end-to-end learning framework such that it can optimize the segmentation and classification networks jointly and iteratively, aiming to not only simplify the training process but also improve the discriminatory power of learned features. Meanwhile, to further improve the accuracy and efficiency of the proposed framework, we will also investigate the automated optimization of the number of selected hard pixels $K$ and weighting factor $\lambda$. Moreover, although our model is built upon the specific application of skin lesion analysis, the proposed mutual bootstrapping strategy itself is generic and can be applied to other deep learning-based medical image segmentation and classification tasks in our future work to improve the performance on both tasks simultaneously.}
	
	\vspace{-0.1cm}
	{
		\bibliographystyle{IEEEtran}
		\bibliography{egbib}
	}
	
\end{document}